%% file: main.tex
\documentclass[journal]{IEEEtran}

\usepackage[utf8]{inputenc} % allow utf-8 input
\usepackage[T1]{fontenc}    % use 8-bit T1 fonts
\usepackage{hyperref}       % hyperlinks
\usepackage{url}            % simple URL typesetting
\usepackage{booktabs}       % professional-quality tables
\usepackage{amsfonts}       % blackboard math symbols
\usepackage{nicefrac}       % compact symbols for 1/2, etc.
\usepackage{microtype}      % microtypography
\usepackage{xcolor}         % colors
\usepackage{amsmath, amssymb, graphicx}
\usepackage{algorithm}
\usepackage{algpseudocode}
\usepackage{wrapfig}
\usepackage{lipsum}
\usepackage{multirow}
\usepackage{cite}
\usepackage{xcolor}
\usepackage{subfig}
\usepackage{comment}

\usepackage{todonotes}

\begin{document}
%\title{Robust and memory-efficient image recovery using monotone operator learning (MOL)}

\title{A Fast, Scalable, and Robust Deep Learning-based Iterative Reconstruction Framework for Accelerated Industrial Cone-beam X-ray Computed Tomography}
\author{Aniket Pramanik,~\IEEEmembership{Member,~IEEE,} Obaidullah Rahman, Singanallur V. Venkatakrishnan, Amirkoushyar Ziabari
        % <-this % stops a space
\thanks{Aniket Pramanik, Obaidullah Rahman, Singanallur V. Venkatakrishnan and Amirkoushyar Ziabari are from Oak Ridge National Laboratory, Oak Ridge, TN, USA 37830 (e-mail: pramanika@ornl.gov, rahmano@ornl.gov, venkatakrisv@ornl.gov, ziabariak@ornl.gov).
This manuscript has been authored by UT-Battelle, LLC, under contract DE-AC05-00OR22725 with the US Department of Energy (DOE). Research sponsored by the US Department of Energy, Office of Energy Efficiency and Renewable Energy (EERE), Advanced Materials \& Manufacturing Technologies Office (AMMTO) and Technology Commercialization Fund (TCF-21-24881), under contract DE-AC05-00OR22725 with UT-Battelle, LLC.
The US government retains and the publisher, by accepting the article for publication, acknowledges that the US government retains a nonexclusive, paid-up, irrevocable, worldwide license to publish or reproduce the published form of this manuscript, or allow others to do so, for US government purposes. DOE will provide public access to these results of federally sponsored research in accordance with the DOE Public Access Plan (http://energy.gov/downloads/doe-public-access-plan).}}
% stops a space

\markboth{Journal of \LaTeX\ Class Files,~Vol.~14, No.~8, August~2021}%
{Shell \MakeLowercase{\textit{et al.}}: A Sample Article Using IEEEtran.cls for IEEE Journals}


\maketitle

\begin{abstract}
  % The abstract paragraph should be indented \nicefrac{1}{2}~inch (3~picas) on
  % both the left- and right-hand margins. Use 10~point type, with a vertical
  % spacing (leading) of 11~points.  The word \textbf{Abstract} must be centered,
  % bold, and in point size 12. Two line spaces precede the abstract. The abstract
  % must be limited to one paragraph.

Cone-beam X-ray Computed Tomography (XCT) with large detectors and corresponding large-scale 3D reconstruction plays a pivotal role in micron-scale characterization of materials and parts across various industries. 
In this work, we present a novel deep neural network-based iterative algorithm that integrates an artifact reduction-trained CNN as a prior model with automated regularization parameter selection, tailored for large-scale industrial cone-beam XCT data. 
Our method achieves high-quality 3D reconstructions even for extremely dense thick metal parts - which traditionally pose challenges to industrial CT images - in just a few iterations. 
Furthermore, we show the generalizability of our approach to out-of-distribution scans obtained under diverse scanning conditions. 
Our method effectively handles significant noise and streak artifacts, surpassing state-of-the-art supervised learning methods trained on the same data.
\end{abstract}

\input{sec/intro}

\input{sec/met}

\input{sec/imp}

\input{sec/exp}

\input{sec/con}

\bibliographystyle{IEEEtran}
\bibliography{main}

%%%%%%%%%%%%%%%%%%%%%%%%%%%%%%%%%%%%%%%%%%%%%%%%%%%%%%%%%%%%

%\appendix

% \section{Appendix / supplemental material}

% Optionally include supplemental material (complete proofs, additional experiments and plots) in appendix.
% All such materials \textbf{SHOULD be included in the main submission.}

\end{document}

%% file: sec/intro.tex
\section{Introduction}
\label{sec:intro} 
X-ray Computed tomography (XCT) is a crucial imaging technique for a wide variety of applications in medicine, science and industry.
It involves solving an inverse problem to reconstruct high-quality images from a collection of measurements from  various angles (views) typically obtained using a cone-beam XCT scanner.  
With recent advances in manufacturing it has been recognized that if one can obtain high-quality reconstruction from fast XCT scans, involving a sparse set of measurements, it can revolutionize the non-destructive characterization of complex parts in industrial settings \cite{ziabari2023enabling,rupal2020geometric, zanini2017assembly, khosravani2020use, naresh2020use}.
A high-throughput pipeline can enable several thousands of parts to be evaluated for the manufacturing defect analysis thereby moving XCT from a tool from one-off analysis to a tool for in-line inspection. 
Industrial XCT scanners used for non-destructive evaluation purposes often have large detectors of size $\approx 2000 \times 2000$ pixels and the reconstruction volumes are $1800 \times 1800 \times 1800$ or larger.
The analytical algorithm Feldkamp-Davis-Kreiss (FDK) \cite{feldkamp1984practical} is a fast reconstruction technique that is most commonly used in commercial industrial XCT systems. 
However, the image quality heavily depends on factors like material properties, desired reconstruction resolution, total integration time (defined as  integration time per view times number of image averaging per view), total number of views and X-ray scan setting (Voltage, Current, physical filters, .
The requirement of long measurement time for dense-metal parts made of important materials (e.g. alloys made of steel) makes FDK impractical for high-throughput industrial XCT. 

Model-based reconstruction techniques have been introduced to estimate the high-quality image from sparsely sampled data.
Model-based Iterative Reconstruction (MBIR) is a class of algorithms that solves the inverse problem by formulating it as an optimization problem consisting of a physics-based model exploiting the measurement domain information and a handcrafted model utilizing some prior information about the image to be reconstructed \cite{venkatakrishnan2021algorithm,yu2010fast,bouman1993generalized,sabo92tv,thibault2007three,venkatakrishnan2021algorithm}.
In the context of industrial XCT, MBIR is not a routine choice because the reconstruction time for the large 3D volumes can be an order of magnitude higher than the scan time. 
Furthermore, choosing useful regularization parameters for MBIR, often done by sweeping parameters in practice, is not feasible for the large-scale reconstruction problems in industrial XCT. 
Recently, authors in \cite{ziabari2023enabling,rahman2023deep} have introduced a single-step supervised deep learning (DL) method that reduces reconstruction time significantly as compared to MBIR. 
It involves a convolutional neural network (CNN) that learns mapping a low-quality FDK reconstruction to a high-quality MBIR reconstruction and the method has shown appreciable performance at high sparsity factors \cite{ziabari2023enabling,rahman2023deep} when compared to MBIR. 
However, this approach requires sufficient training data from a variety of acquisition settings which is often unavailable and that leads to lack of generalizability on out-of-distribution data encountered during variation in acquisition conditions.
In order to mitigate this, the CNN would require re-training whenever a new setting is encountered, which could in turn impact their wider adaptability to inspection in manufacturing industries.

Several algorithmic developments have been made for other imaging applications such as magnetic resonance imaging (MRI) and medical XCT where DL-based learned models have been combined with physics-based models. 
These methods have reduced computational complexity as compared to MBIR and have improved performance over single-step DL \cite{meinhardt2017learning, dong2018denoising,zhang2017learning, jin2017deep, kang2017deep, chen2017low, mccann2017convolutional,lucas2018using,ronneberger2015u,jinUnet17} with limited availability of training data.
The improvement in generalizability over single-step DL is achieved by enforcing the solution to be consistent with the associated physics-based forward model and the measurements obtained from the scanner.
These approaches can be broadly classified into Loop Unrolled (LU) \cite{adler2018learned, moriakov2023end, rudzusika20243d,monga2021algorithm, zhang2018ista,pramanik2020deep} Deep Learning and Plug-and-Play (PnP) methods  \cite{venkatakrishnan2013plug, sreehari2016plug, zhang2019deep,zhang2021plug,kamilov2023plug,chan2016plug, sun2019online, ryu2019plug, liu2021recovery,romano2017little,reehorst2018regularization}. 
The PnP framework integrates a pre-trained CNN-based Gaussian denoiser as a prior model into a convergent MBIR algorithm.
High computational complexity for cone-beam XCT has hindered the use of PnP in industrial settings despite of researchers showing the potential of the approach to improve image quality \cite{majee2021multi}.
LU instead unrolls a CNN regularized optimization algorithm for a pre-defined number of outer iterations (typically $\leq 10$) for end-to-end training.
Due to fewer outer iterations (when compared to MBIR and PnP) and all the gradient-based computations as well as storage happening in GPU (graphical processing unit), LUs are significantly faster. 
However, the high GPU memory requirement challenges its scalability to practical industrial cone-beam CT.

\begin{figure}
\centering
\includegraphics [scale=0.4,keepaspectratio=true]
{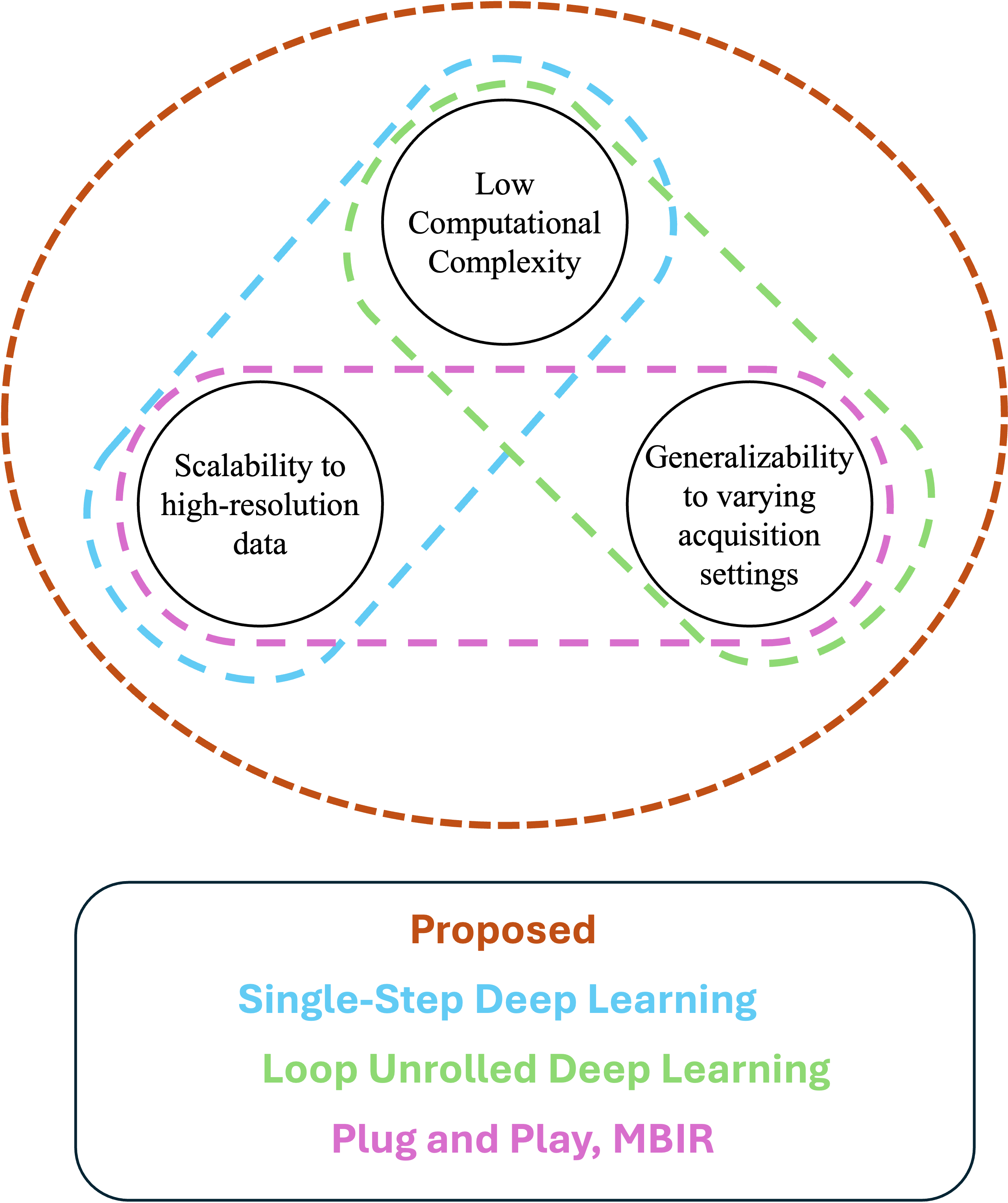}
\caption{Desired properties of computed tomography reconstruction algorithms required for high-quality imaging of large 3D volumes encountered in industrial CT settings.
Existing approaches in the literature satisfy two of the three properties.
}
\label{fig:tri}
\end{figure}

In the context of cone-beam XCT, recent developments have been on low computation and memory-efficient LU/PnP methods \cite{lahiri2023sparse, moriakov2023end, rudzusika20243d}.
However, these approaches have been validated on reconstruction volumes ($ \approx 500 \times 500 \times 500$) that are significantly smaller than practical industrial XCT.
While these developments have taken a stride towards addressing some of the key challenges mentioned in Fig. \ref{fig:tri} including (i) low computational complexity, (ii) scalability to large datasets and (iii) generalizability, the methods do not target practical industrial cone-beam XCT.
The scalability property in Fig. \ref{fig:tri} refers to efficiency in GPU memory consumption when the algorithm is applied to large-scale image data ($\geq 2000^3$ volume sizes) ensuring the demand does not exceed the existing standard of computing resources.
The generalizability property (Fig. \ref{fig:tri}) highlights an algorithms's robustness in terms of reconstruction quality for data obtained from acquisition conditions not included in the training set of the CNN, and are therefore, unseen by the network.
Single-step supervised DL methods have been shown to be underperforming in such conditions.

In this paper, the goal of our research is to provide a practical algorithm that satisfies the desired properties in Fig. \ref{fig:tri} and therefore facilitate high-quality reconstruction of industrial XCT data.
We introduce a CNN-regularized physics-based inversion recovery algorithm based on a PnP framework that is inspired from the Half-Quadratic Splitting (HQS) formulation. 
The proposed method approximates the proximal operation associated with the prior model using a pre-trained CNN and performs the regularized inversion using the conjugate gradient (CG) algorithm.
The number of alternations between the CNN and CG steps are chosen as per the desired reconstruction quality.
The pre-trained CNN is an artifact removal network that learns to suppress artifacts from low-quality sparse-view FDK reconstructions and the training is memory-efficient due to the paired data being small 2D patches extracted from large-scale 3D images.
Additionally, we propose a regularization parameter selection strategy for the CG algorithm. 
It ensures that the regularization strength is automatically adjusted according to the amount of noise/artifacts present in the reconstruction thereby addressing a critical challenge when dealing with CT reconstruction algorithms in practice. 
In a demonstration of generalizability of the proposed algorithm, we compare the method against existing methods showing impressive performance on XCT scans with varying X-ray source voltage, total integration time and sparsity. 
The regularization selection strategy shows better performance as compared to having a fixed regularization parameter. 
In addition, it has significantly lower computational complexity compared to MBIR due to the lower number of outer iterations.  
Although the proposed algorithm is tailored for cone-beam XCT applications, it is general enough for its extension to other large-scale image reconstruction applications.

While the proposed method uses an artifact-removal CNN as a prior, some recent works \cite{hurestoration, liu2020rare} have also explored DL-based artifact removal/restoration priors instead of Gaussian denoising in a PnP/RED (Regularization by Denoising) framework.
A key difference of our method with \cite{liu2020rare} is the utilization of an implicit prior instead of an explicit one obtained from the RED formulation.
In addition, we perform a supervised training of the CNN for improved performance instead of the self-supervised approach in \cite{liu2020rare}.
The other work \cite{hurestoration} employs deblurring/super-resolution restoration as priors irrespective of the measurement model.
Although this approach provides convergence guarantees, the mismatch in the type of artifacts due to measurement and prior models can lower performance as shown in the case of Gaussian denoising (see supplementary section in \cite{hurestoration}).
We instead train the CNN to reduce acquisition specific artifacts arising from a combination of factors including beam-hardening and undersampling.

%% file: sec/met.tex
\section{The Algorithm}
\begin{comment}This work mainly focuses on mitigating three key challenges faced by existing industrial XCT reconstruction algorithms: scalability to large-scale 3D image data, reduced computational complexity and robustness in performance (generalizability) on data acquired from varying acquisition conditions in an industrial setup.
We address the above-mentioned by (1) proposing an iterative algorithm for a model-based image recovery using a data-consistency block; (2) a patch-based 2D UNET as a prior for reduced GPU memory demand and (3) an iteration-dependent regularization parameter selection strategy for faster convergence to the desired solution.
\end{comment}

\begin{figure}%[t!]
\centering
\includegraphics [scale=.4]{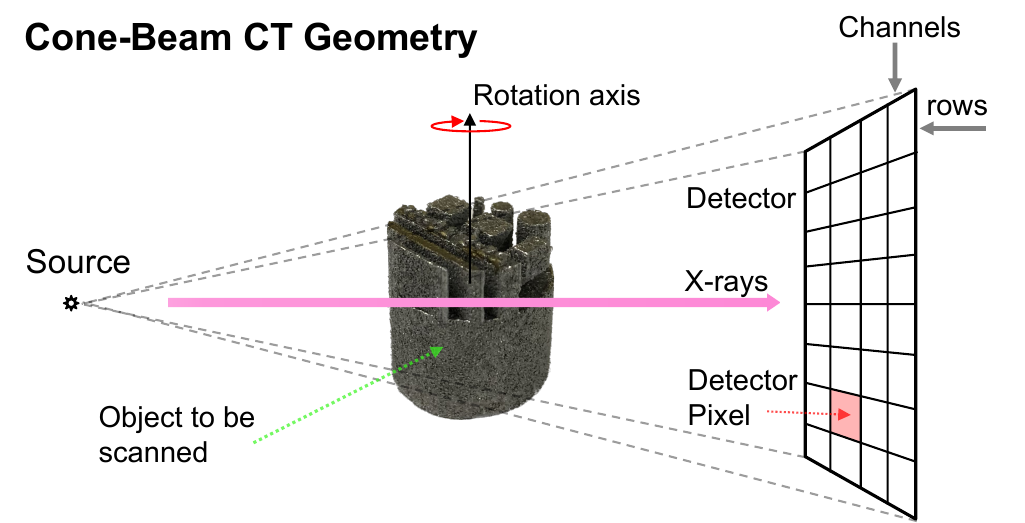}
\caption{Schematic of a cone-beam X-ray computed tomography with common nomenclature.}
\label{CBCT}
%\vspace{-0.5cm}
\end{figure}

\subsection{Model-based Recovery Using Half-Quadratic Splitting}
An industrial XCT system comprises a cone-beam X-ray source interacting with a target and a detector measuring the transmitted X-rays (see Fig.~\ref{CBCT}), also termed as a view or projection. 
To perform a CT scan, the manufactured part is rotated about a single axis, acquiring multiple measurements. 
A common model for measurements obtained from an industrial XCT system is $\mathbf{y} = \mathbf A \mathbf x$,
where vector $\mathbf{y} \in \mathbb R^{\rm M}$ denotes the log-normalized projection measurements, $\mathbf A \in \mathbb R^{\rm M \times N}$ is the linear operator representing cone-beam forward model and vector $\mathbf x \in \mathbb R^{\rm N}$ is the underlying 3D image volume of linear attenuation coefficients to be reconstructed.
%The dimension $\rm M = r \times c \times v$ where $\rm r$, $\rm c$ and $\rm v$ denote the number of detector rows, columns and views respectively and for the underlying reconstructed image, dimension is represented by $\rm N = d \times h \times w$ where $\rm d$, $\rm h$ and $\rm w$ are depth, height and width of the 3D image.

A state-of-the-art method to reconstruct the 3D volume $\mathbf x$ is by using a regularized weighted least-squares formulation, also referred to as model-based iterative reconstruction (MBIR), and is  given by 
\begin{eqnarray}
%\mathbf x_{\rm MAP} & = & \arg \min_{\mathbf x} -\log p(\mathbf y| \mathbf x) - \log p(\mathbf x) \\
\hat{\mathbf{x}}& = & \arg \min_{\mathbf x} \underbrace{\frac{1}{2}\|\mathbf A \mathbf x - \mathbf y\|_{2}^2}_{\text{Physics-based Model}} + \lambda \underbrace{\mathbf R(\mathbf x)}_{\text{Regularizer}} 
\label{reg}
\end{eqnarray}
where $\lambda$, the regularization parameter, balances the effect of the two terms. 
The first term is the model encoding the acquisition physics and the second term $\mathbf R(\mathbf x)$ is a regularizer that ensures certain desirable properties in the reconstruction.
The MBIR formulation in \eqref{reg} for cone-beam XCT considers unity weight for the data-consistency term.

Instead of precisely solving an optimization algorithm, a half-quadratic splitting of \eqref{reg} results in an iterative algorithm alternating between the following sub-problems, 
\begin{eqnarray}
\mathbf z_k = \arg \min_{\mathbf z} \lambda \mathbf R(\mathbf z) + \frac{\beta}{2} \|\mathbf x_{k-1} - \mathbf z\|_2^2 = \rm prox_{\frac{\lambda}{\beta}}(\mathbf x_{k-1}) \label{pr} \\
\mathbf x_k = \arg \min_{\mathbf x} \frac{1}{2}\|\mathbf A \mathbf x - \mathbf y\|_2^2 + \frac{\beta}{2}\|\mathbf x - \mathbf z_k\|_2^2. \label{img}
\end{eqnarray}
where $\rm prox(\cdot)$ is a proximal map operator. 
We note that while ideally, the value of $\beta$ has to be increased gradually for the algorithm to potentially converge to the solution of the original optimization problem, in practice the above HQS method produces one solution that depends on the value of $\beta$.
The sub-problem in \eqref{img} is a quadratic cost function which solves for $\mathbf x$ using, $\mathbf x_k = (\mathbf A^T \mathbf A + \beta \mathbf I)^{-1}(\mathbf A^T \mathbf y + \beta \mathbf z_k)$ where $\mathbf A^T$ is a transpose of $\mathbf A$. 
For industrial cone-beam XCT, implementing the regularized inversion step $(\beta \mathbf I + \mathbf A^T \mathbf A )^{-1}$ is non-trivial due to the large sizes of the matrices and therefore, \eqref{img} is solved approximately using either the conjugate gradient (CG) or gradient descent (GD) based algorithms.
The proximal operation in \eqref{pr} is a maximum aposteriori (MAP) estimate of a Gaussian denoising problem \cite{kamilov2023plug} which is approximated using a convolutional neural network for improved performance and speed.

\subsection{Proposed Iterative Algorithm}
Instead of a CNN-based Gaussian denoiser for \eqref{pr}, we utilize an artifact removal CNN to replace the proximal operator.
We pre-train a CNN for removing artifacts from a sparse-view FDK reconstruction and that is plugged as a regularizer into the sub-problem \eqref{img}.
The artifacts in an initial FDK reconstruction are due to degradation occurring from a combination of factors including beam-hardening, sparse-views and low signal-to-noise ratio data. 
Depending on the material property (density), the amount of artifacts can vary from mild for lower density material to severe for highly dense materials.
Scan setting (choice of pre-filter, integration time, number of image averaging per image, voltage, current) directly impacts the quality of the scans (such as beam hardening and SNR values).
On top of that, sparsity introduces streak artifacts in the images.
%We hypothesise that a standard Gaussian denoising CNN might not be ideal for the removal of such artifacts and therefore, train an artifact removal CNN instead.

Unlike conventional PnP algorithms, we introduce a regularization parameter selection strategy for $\beta$ in  \eqref{img}.
We keep the regularization parameter adaptive for improved performance as well as reduced computational complexity of the algorithm from fewer outer iterations.
The proposed algorithm alternates between the following three steps,
\begin{eqnarray}
\mathbf z_k &\leftarrow& \mathbf D_{\theta}(\mathbf x_{k-1}) \label{cnn} \\
\beta_k &\leftarrow& \textit{RegularizationSelection}(\mathbf z_k, \mathbf A_c, \mathbf y_c) \label{regsel} \\
\mathbf x_k &\leftarrow& \arg \min_{\mathbf x} \frac{1}{2}\|\mathbf A \mathbf x - \mathbf y\|_2^2 + \frac{\beta_k}{2}\|\mathbf x - \mathbf z_k\|_2^2. 
\label{vimg}
\end{eqnarray}
The pre-trained CNN is denoted by $\mathbf D_{\theta}(\cdot): \mathbb R^{\rm N} \to \mathbb R^{\rm N}$ with trainable parameters $\theta$ and it operates in spatial domain. 
The image update sub-problem in \eqref{vimg} uses the parameter $\beta_k$ selected for the iteration $k$ in  \eqref{regsel}. 
The pseudo-code of the algorithm is shown in Algorithm \ref{algo:inference}.
The algorithm is initialized with an FDK reconstruction and it is run for a pre-defined number of iterations $K$.
The data-consistency step in \eqref{vimg} comprises of the CG algorithm.
We discuss the CNN architecture and the regularization selection step in more details in the subsequent sections.

\begin{algorithm}[t!]
  \caption{Proposed Algorithm}
  \label{algo:inference}
  \begin{algorithmic}
  \State \textbf{Input:} low-quality image $\mathbf x_0 = \mathbf x_{\rm FDK}$, $\mathbf A$, $\mathbf A_c$, $\mathbf y$, $K$
  \State \textbf{Output:} Restored Image $\widehat{\mathbf x}$
  \State $\mathbf y_c \leftarrow$ Measurements from center rows of $\mathbf y$.
\While{$k \leq K$}
    \State $\mathbf z_k \leftarrow \mathbf D_{\theta}(\mathbf x_{k-1})$
    \State $\beta_k \leftarrow$ \textit{RegularizationSelection}($\mathbf z_k$, $\mathbf A_c$, $\mathbf y_c$)
    \State $\mathbf x_k \leftarrow \arg \displaystyle\min_{\mathbf x} \frac{1}{2}\|\mathbf A \mathbf x - \mathbf y\|_2^2 + \frac{\beta_k}{2} \|\mathbf x - \mathbf z_k\|_2^2$
    \State (Conjugate Gradient Algorithm)
    
\EndWhile \\
\Return $\widehat{\mathbf x} = \mathbf x_K$
\end{algorithmic}
\end{algorithm}

\subsection{Regularization Parameter Selection Strategy}
Typically, PnP algorithms keep the regularization parameter $\beta$ fixed for successive outer iterations since it corresponds to a principled approach to finding a consensus equilibrium \cite{buzzard2018consensus}.  
Instead, we propose to choose a different value for the parameter $\beta$ at every iteration.
Specifically, we run a grid search by reconstructing only a few center slices from the 3D volume and evaluate the reconstruction quality using a reference-free image quality metric. 
Since we only reconstruct a few center slices, the regularization selection is extremely fast and the overall selection time is significantly minimal compared to that of the whole 3D image reconstruction. 
The proposed regularization selection algorithm is shown in Algorithm \ref{algo:reg}.

The function $RegularizationSelection(\cdot)$ has arguments comprising of the cone-beam XCT operator $\mathbf A_c$ for center slices, the corresponding measurements $\mathbf y_c$ extracted from center detector rows of the full set of measurements $\mathbf y$ and the CNN reconstruction $\mathbf r$.
%The dimensions are ${\rm N_c = rc \times h \times w}$ and ${\rm M_c = rc \times c \times v}$ where $\rm h,w,c,v$ are height, width, detector columns and views respectively and are similar to the ones defined for $\mathbf A$. 
We consider a set of $n$ values of $\beta$ lying on a geometric progression with common ratio $r$ and the first term $a$.
The set is given as $\mu = \{\mu_1, \mu_2, \cdots, \mu_n\}$ where $\mu_i = a*r^{i-1}$.
We choose $\beta = \mu_i$ for $i = \arg \min_{i} \text{BRISQUE}(\mathbf v_i)$ where the index $i$ corresponds to the center slice reconstruction with the lowest BRISQUE score \cite{mittal2012no}.

BRISQUE score \cite{mittal2012no} is an unsupervised performance metric for assessing the reconstruction quality for each $\mu_i$. 
It is a score generated by a pre-trained neural network based on extracted features of the input image.  
BRISQUE score has been shown to be an effective tool to characterize images based on their perceptual quality (noise content and sharpness). 
It is a score between 0-100 with the scores being low for an image with low noise content and sharper edges.
A higher BRISQUE score implies lower image quality.
The computational cost associated with sweeping $\beta$ at each iteration is orders of magnitude lower compared to the outer iteration run-time. 

\begin{algorithm}[t!]
  \caption{$RegularizationSelection$}
  \label{algo:reg}
  \begin{algorithmic}
  \State \textbf{Param.:} init. $r = 0.5$, $a = 2$, $n = 14$, $q = 100$ 
  \State \textbf{Input:} CNN output $\mathbf r$, $\mathbf A_c$, $\mathbf y_c$ 
  \State \textbf{Output:} Regularization Parameter $\beta$
  \State $\mathbf r_c \leftarrow$ Extract center slices from $\mathbf r$.
  \For{$i = 1, \ldots , n$}
  \State $\mu_i = a*r^{i-1}$
  \State $\mathbf v_i \leftarrow \arg \displaystyle\min_{\mathbf v} \frac{1}{2}\|\mathbf A_c \mathbf v - \mathbf y_c\|_2^2 + \frac{\mu_i}{2} \|\mathbf v - \mathbf r_c\|_2^2$  
  \State (Conjugate Gradient Algorithm)
  % \State $cbq \leftarrow \text{BRISQUE}(\mathbf v_i)$
  % \If{$cbq < bq$}
  \If{$\text{BRISQUE}(\mathbf v_i) < q$}
  \State $q = \text{BRISQUE}(\mathbf v_i)$
  \State $\beta = \mu_i$
  \EndIf
  \EndFor \\
\Return $\beta$
\end{algorithmic}
\end{algorithm}

\subsection{The Overall Workflow}
\begin{figure*}[h!]%
\centering
\subfloat[Beam Hardening Correction]{%
\includegraphics[scale=0.42,keepaspectratio=true]{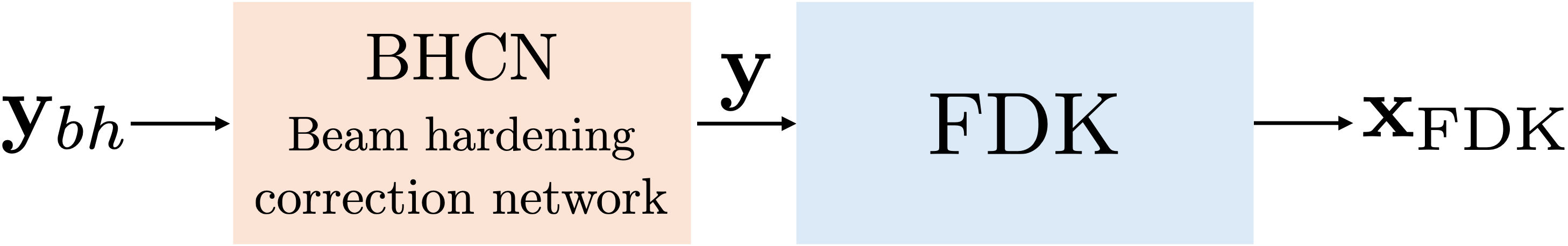}
}

\vspace{15pt}
\subfloat[Proposed Iterative Algorithm]{
\includegraphics[scale=0.42,keepaspectratio=true]{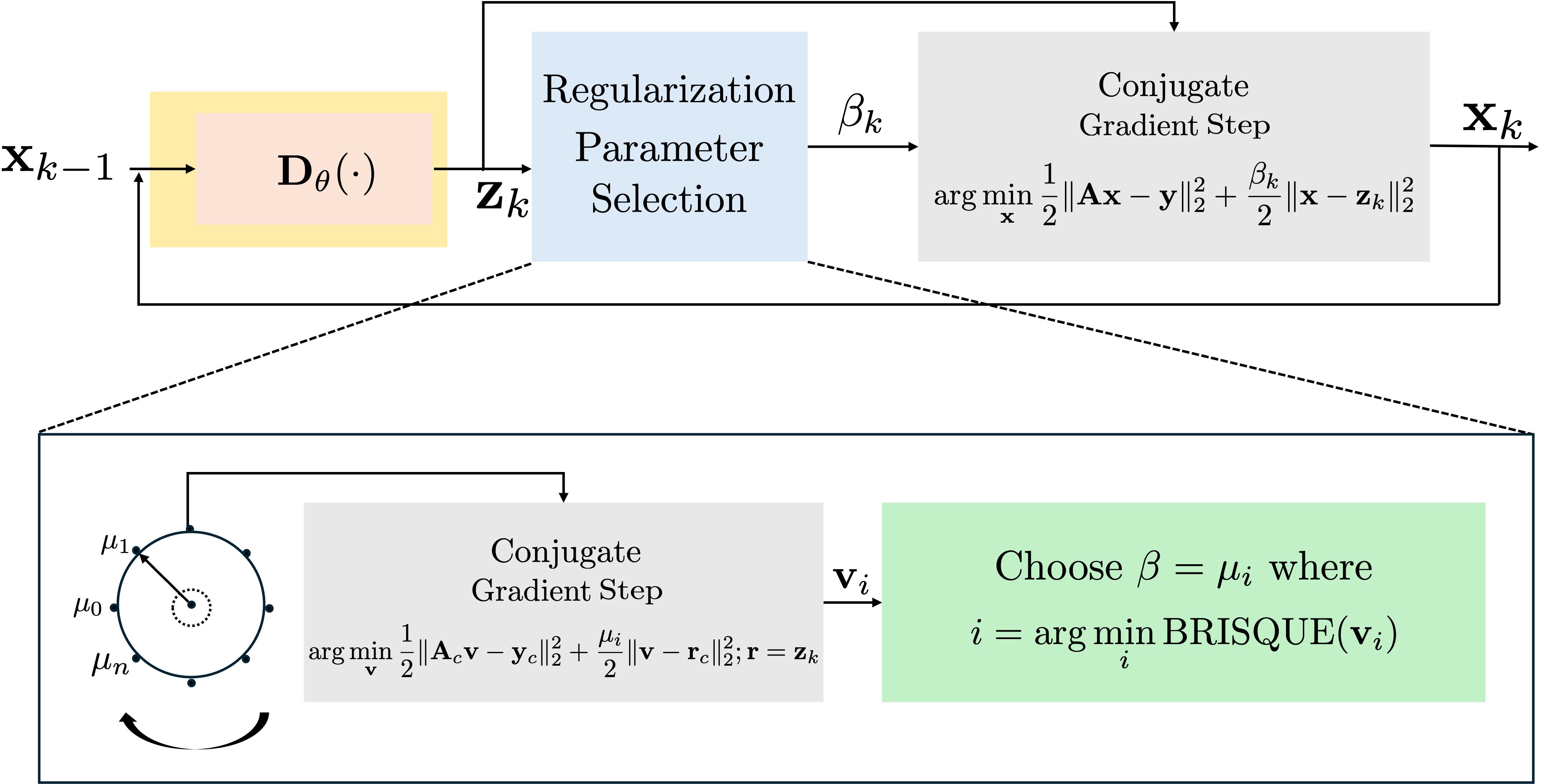}
}
\caption{Illustration of the proposed algorithm. 
Our method alternates between an artifact removal CNN and a few iterations of a conjugate gradient algorithm that enforces data-consistency while ensuring the result is close to the output of the CNN. 
Additionally in each outer iteration, the regularization parameter for the CG stage is adjusted based on as assessment of image quality. 
This architecture ensures a high-quality reconstruction with few outer iterations while being more generalizable compared to vanilla single-step CNNs.}%
\label{fig:arch}%
\end{figure*}

We initialize the proposed algorithm with an FDK reconstruction on pre-corrected measurements in order to handle artifacts from beam hardening which is common when a high-density material is scanned. 
The workflow is shown in Fig. \ref{fig:arch}.
We apply a pre-trained neural network \cite{rahman2023deep} on the projection data $\mathbf y_{bh}$ (measurements) to obtain beam-hardening corrected projections $\mathbf y$.  
%It is a Multi-layer Perceptron (MLP) that generates parameters for the Van de Casteel model \cite{van2002energy} of beam hardening.
Therefore, the workflow involves beam hardening correction in the measurement domain followed by a fast reconstruction using the FDK algorithm that is fed to the proposed iterative reconstruction algorithm.
The number of CG iterations for the image update step and the regularization selection step is chosen as ten based on an ablation study.
We have discussed the impact of CG iterations in Fig. \ref{fig:cgb} as well as the results section.

\subsection{Space Complexity and Memory Efficient Implementation}
If we consider $K$ as number of outer iterations of an iterative algorithm, the space complexity of LU architecture is $\mathcal O(K)$ since it requires unrolling the network for $K$ iterations for end-to-end training of the algorithm through backpropagation \cite{adler2018learned, moriakov2023end, rudzusika20243d,monga2021algorithm, zhang2018ista,pramanik2020deep}.
Single-step supervised DL \cite{ziabari2023enabling,rahman2023deep} and PnP methods \cite{venkatakrishnan2013plug, sreehari2016plug, zhang2019deep,zhang2021plug,kamilov2023plug,chan2016plug, sun2019online, ryu2019plug, liu2021recovery,romano2017little,reehorst2018regularization} instead require space for training a single CNN and therefore the complexity is $\mathcal O(1)$.
Similar to PnP, the proposed iterative DL method pre-trains a CNN that is plugged into an iterative algorithm and therefore reduces the complexity of LU by a factor of $K$ ($\mathcal O(1)$). 

For DL-based methods, operating a 2D or 3D CNN on the reconstruction volume of size $1800 ^ 3$ (approximately) is memory demanding.
Therefore, we split the images into 2D patches $(256 \times 256)$ during training to have sufficient GPU memory for storing the backpropagation gradients.
The splitting of data into patches allows memory efficient training and inference using CNNs irrespective of the size of the reconstruction volume, making it scalable for very large datasets.

%% file: sec/imp.tex
\section{Implementation Details}
\label{sec:impd}

\subsection{Datasets}
Experiments were performed on cone-beam XCT scans of 3D printed parts made of stainless steel.
The scans were performed using a Carl Zeiss Metrotom 800 CT scanner with a $1840\times1450$ detector with a pixel pitch of 127$\rm \mu m$.
The scan details of all the datasets are shown in Table \ref{tab:ct_data}.
The part geometry is similar for all the scans.
For training the CNN, we used paired data from scans of three different parts and the acquisition settings were 200 kV source voltage with 8s total integration time per view (1s of integration time times 8 image averaging for better SNR).
The paired data consists of 2D patches extracted from 3D image volumes of sparse-view FDK as input and the corresponding dense-view MBIR as ground truth. 
The number of sparsely sampled views for FDK were 138 in a short-scan setting ranging from $0^{\circ}-(180^{\circ}$ + \text{fan angle})~\cite{parker1982optimal,ziabari2018model}.
Given that the detector has 1840 channels, and considering Nyquist criteria for spatial resolution in XCT, this is considered extremely sparse input data by a factor of $13\times$.
The ground truth dense-view MBIR is obtained from a long-scan consisting of measurements from 1000 views ranging from $0^{\circ}-360^{\circ}$. 

The data for inference was obtained by scanning three different steel samples denoted as ST, ST-1 and ST-2 respectively.
We scanned the ST sample for nine different settings including combinations of three different X-ray peak source voltages (160 kV, 180 kV, 200 kV) and three total integration times (0.6 seconds, 1.8 seconds, 3.6 seconds) leading to total scan times of 10 minutes, 30 minutes and 60 minutes, respectively, for 1000 views in each case. 
All the acquisition settings for the sample ST are out-of-distribution (OOD) for the CNN.
Decreasing the voltage (power), will result in reduced transmission through the object and in turn lower SNR on the projection data, which in turn impact the quality of the reconstruction and noise variances.
Further, reducing the scan time through reducing the number of views as well as total integration time per each view, directly results in various low SNR projections and reconstructions with noise variances and texture that are different from training data.
This allows us to test our algorithm's performance and generalizability on varying scenarios under OOD data.
Sample ST-1 was a long-scan ($0^{\circ}-360^{\circ}$) with 1066 views and the sparsely sampled views (74) were obtained through a reduction by a factor of eight from the corresponding short-scan ($0^{\circ}-(180 + \text{fan angle})^{\circ}$). 
For ST-2 we performed a short-scan with 580 views and that was sub-sampled by a factor of eight to obtain 73 views.
Since the training inputs were 138 views short-scan, the sparsity levels of ST-1 and ST-2 with only 73-view short-scan are larger, and therefore, this is an OOD setting for our CNN.
The goal is to test our algorithm for a different sparsity factor in ST-1 and ST-2.
We also test our algorithm on parts made of AlCe (ALC) and Inconel-718 (here we call it IN) as described in Table \ref{tab:ct_data}.
The part geometry is similar to the steel scans ST.
AlCe has lower density and typically produce reconstructions with less noise and beam hardening. 
For a supervised learning approach, this pose a challenge as it tries to over-correct the noise and/or over-correct beam hardening (smooth/saturate the reconstruction).
On the other hand, Inconel-718 is significantly denser than steel and therefore the amount of noise and beam hardening is larger in a typical FDK reconstruction.
A single-step supervised learning approach will face challenges to deal with the noise variance and texture that is OOD with respect to its training data.
We will evaluate and report performance of our algorithm on these OOD data as well.

\begin{table*}[ht!]
\fontsize{8}{11}
\selectfont
\centering
\begin{tabular}{|c|cccccccc|}
\hline
\multicolumn{9}{|c|}{Cone-beam X-ray CT Data} \\ \hline
\multirow{3}{*}{Dataset} & \multirow{3}{*}{Sample} & \multirow{3}{*}{Material} & {Approximate} & \# of views & Short-Scan & Source  & Total & Scan Time\\ 
& & & Density & [Input, Ref.] & [Input, Ref.] & Voltage & Integration & [Input, Ref.] \\ 
& & & ($gr/cm^3$) & & & & Time & (minutes) \\ \hline 
\multirow{3}{*}{Training}  &  & Stainless Steel &  7 &  [138, 1000] & [True, False] & 200 kV & 8s & [18.4, 133]\\
 & & Stainless Steel  & 7 & [138, 1000] & [True, False] & 200 kV & 8s & [18.4, 133] \\
 & & Stainless Steel  & 7 & [138, 1000] & [True, False] & 200 kV & 8s & [18.4, 133] \\ \hline
\multirow{9}{*}{Testing} & ST & Stainless Steel & 7 & [1000, -] & [True, -] & 160 kV & 0.6s & [10, -]\\
 & ST & Stainless Steel & 7 & [1000, -] & [True, -] & 160 kV & 1.8s & [30, -]\\
 & ST & Stainless Steel & 7 & [1000, -] & [True, -] & 160 kV & 3.6s & [60, -]\\
 & ST & Stainless Steel & 7 & [1000, -] & [True, -] & 180 kV & 0.6s & [10, -]\\
 & ST & Stainless Steel & 7 & [1000, -] & [True, -] & 180 kV & 1.8s & [30, -]\\
 & ST & Stainless Steel & 7 & [1000, -] & [True, -] & 180 kV & 3.6s & [60, -]\\
 & ST & Stainless Steel & 7 & [1000, -] & [True, -] & 200 kV & 0.6s & [10, -]\\
 & ST & Stainless Steel & 7 & [1000, -] & [True, -] & 200 kV & 1.8s & [30, -]\\
 & ST & Stainless Steel & 7 & [1000, -] & [True, -] & 200 kV & 3.6s & [60, -]\\
 & ST-1 & Stainless Steel & 7 & [73, 580] & [True, True] & 200 kV & 8s & [10, 78]\\
 & ST-2 & Stainless Steel & 7 & [74, 1066] & [True, False] & 200 kV & 8s  & [10, 156]\\
& ALC & AlCe & 3.4 & [145, 580] & [True, True] & 180 kV & 4s & [10, 39]\\
& IN & Inconel 718 & 8.2 & [143, 2132] & [True, False] & 220 kV & 4s & [10, 142]\\
 \hline
\end{tabular}
\vspace{1em}
\caption{Cone-beam XCT Data Description. 
Increasing the total integration time per view improves the quality of the raw data and make the reconstruction less noisy.
Higher source voltages provides improved measurements and higher sparsity factor increases artifacts in the images.
Total integration time is the multiplication of the integration time per image and the number of images averaged per view.
The total scan time is obtained by multiplying total integration time with the number of views.
These data are selected to evaluate our model's generalizability to a variety of out-of-distribution data that are designed to make reconstruction more challenging by reducing the quality of scans (reduced number of views or total integration time or source voltage).
Source current is not provided, but it is adjusted by the user for X-ray to have enough transmission for all the scans.}
\label{tab:ct_data} 
\end{table*}

\subsection{Methods for Comparison}
The proposed method is compared against state-of-the-art algorithms for industrial cone-beam XCT reconstruction including FDK, MBIR \cite{yu2010fast} and DLMBIR \cite{rahman2023deep}. 
We term the methods operating on beam-hardening corrected data as BHCN-X where "X" denotes the method name.
%MBIR \cite{yu2010fast} is an iterative optimization algorithm that regularizes the image recovery using the Gaussian Markov Random Field prior.
DLMBIR  is a single-step DL approach that uses a 2.5D UNET for mapping a low-quality reconstruction to a high-quality reconstruction~\cite{ziabari2023enabling}. 
%a neighbourhood of five 2D slices to the corresponding central slice.
%The 2D slices are stacked along the channel dimension and this approach helps to extract more structural information.
%FDK \cite{feldkamp1984practical} is a linear filtering-based method for analytical inversion of measurements into a 3D volume reconstruction.

\subsection{CNN Architecture, Training and Inference}
We choose a 2D full-size residual UNET \cite{ronneberger2015u} architecture for $\mathbf D_{\theta}(\cdot)$. 
It consists of four pooling/unpooling layers, 3 x 3 convolutional filters with the output channels in the beginning layer being 64.
The UNET is trained in a supervised fashion using the mean squared error loss between patches of low-quality reconstruction and the corresponding ground truth data.
The input data is a sparse-view FDK reconstruction obtained from the beam hardening corrected projection data as shown in Fig. \ref{fig:arch} and the ground truth is the corresponding dense-view MBIR reconstruction.
The model is trained for 100 epochs on patches of size $256 \times 256$ and the UNET \cite{ronneberger2015u} parameters are optimized using Adam \cite{kingma2014adam}.
We use four NVIDIA P100 GPUs for training $\mathbf D_{\theta}$ using data parallelism across batches.
For inference, $\mathbf D_{\theta}$ operates sequentially on the stack of full-size 2D slices from the 3D image volume and the execution is parallelized across GPUs along the third dimension of the 3D data. 
In the block for conjugate gradient step (Fig. \ref{fig:arch}), we use the ASTRA toolbox \cite{van2015astra,palenstijn2011performance,van2016fast} to implement the forward operators ($\mathbf A, \mathbf A_c$) and they are executed in parallel across multiple GPUs. The rest of the operations in CG block are performed in CPU (central processing unit).

\begin{table*}[ht!]
\fontsize{8}{11}
\selectfont
\centering
\begin{tabular}{|cc|cccccc|}
\hline
\multicolumn{8}{|c|}{Variation in Source Voltage (Integration Time: 0.6s)} \\ \hline
 \multirow{2}{*}{Sample} & Source & \multicolumn{2}{c}{BHCN-FDK} & \multicolumn{2}{c}{BHCN-DLMBIR} & \multicolumn{2}{c|} {Proposed}\\ 
& Voltage & PSNR $(\uparrow)$ & SSIM $(\uparrow)$ & PSNR $(\uparrow)$ & SSIM $(\uparrow)$ & PSNR $(\uparrow)$ & SSIM $(\uparrow)$\\ \hline 

ST & 160 kV & 21.13 & 0.152 & 34.93 dB & 0.864 & \textbf{37.52 dB} & \textbf{0.984} \\
ST & 180 kV & 22.13 & 0.182 & 36.32 dB & 0.916 & \textbf{38.32 dB} & \textbf{0.989} \\ 
ST & 200 kV & 24.96 & 0.385 & 38.89 dB & 0.992 & \textbf{38.91 dB} & \textbf{0.992} \\
\hline \hline
\multicolumn{8}{|c|}{Variation in Integration Time (Source Voltage: 200 kV)} \\ \hline
\multirow{2}{*}{Sample} & Integration & \multicolumn{2}{c}{BHCN-FDK} & \multicolumn{2}{c}{BHCN-DLMBIR} & \multicolumn{2}{c|} {Proposed}\\ 
& Time & PSNR $(\uparrow)$ & SSIM $(\uparrow)$ & PSNR $(\uparrow)$ & SSIM $(\uparrow)$ & PSNR $(\uparrow)$ & SSIM $(\uparrow)$\\ \hline 
ST & 0.6s & 23.14 dB & 0.249 & 33.70 dB & 0.949 & \textbf{34.56 dB} & \textbf{0.968} \\
ST & 1.8s & 27.50 dB & 0.432 & 33.82 dB & 0.982 & \textbf{34.96 dB} & \textbf{0.970} \\
ST & 3.6s & 28.98 dB & 0.628 & 32.90 dB & 0.924 & \textbf{35.11 dB} & \textbf{0.986} \\ \hline \hline
\multicolumn{8}{|c|}{Variation in Sparsity} \\ \hline
\multicolumn{2}{|c|}{\multirow{2}{*}{Sample}} & \multicolumn{2}{c}{BHCN-FDK} & \multicolumn{2}{c}{BHCN-DLMBIR} & \multicolumn{2}{c|} {Proposed}\\ 
& & PSNR $(\uparrow)$ & SSIM $(\uparrow)$ & PSNR $(\uparrow)$ & SSIM $(\uparrow)$ & PSNR $(\uparrow)$ & SSIM $(\uparrow)$\\ \hline 
\multicolumn{2}{|c|}{ST-1} & 26.10 dB & 0.458 & 33.06 dB & 0.918 & \textbf{35.50 dB} & \textbf{0.952} \\ 
\multicolumn{2}{|c|}{ST-2} & 28.23 dB & 0.531 & 34.10 dB & 0.928 & \textbf{36.72 dB} & \textbf{0.969} \\ \hline
\end{tabular}
\vspace{1em}
\caption{Reconstruction performance of the proposed algorithm is compared against the state-of-the-art methods in terms of PSNR (in dB) and SSIM at various settings. The top sub-table shows comparisons on the sample ST scanned at different voltages (160 kV, 180 kV) at a fixed integration time of 0.6 seconds (10 minute scan time). For computing metrics, we have considered MBIR \cite{yu2010fast} on ST at (180 kV, 3.6 seconds integration time) as the reference. The middle sub-table reports the performance on ST scanned at different integration times (0.6 seconds, 1.8 seconds, 3.6 seconds) at a source voltage of 200 kV peak. The metrics have been computed using MBIR \cite{yu2010fast} on ST at (200 kV, 3.6 seconds) as reference. The bottom sub-table shows the performance on samples ST-1 and ST-2 that are reconstructed from 73/74 views at an undersampling factor of 8 when considered the views from short-scan. This undersampling factor is higher by a factor of two when compared to that of the training data. The metrics are computed against the MBIR \cite{yu2010fast} reconstruction on reference scans (580/1066 views) for each sample as shown in Table \ref{tab:ct_data}. The metrics of the best performing method are highlighted in bold for each case.}
\label{tab:metrics} 
\end{table*}

\begin{figure*}[h!]%
\centering
\includegraphics[scale=0.4,keepaspectratio=true]{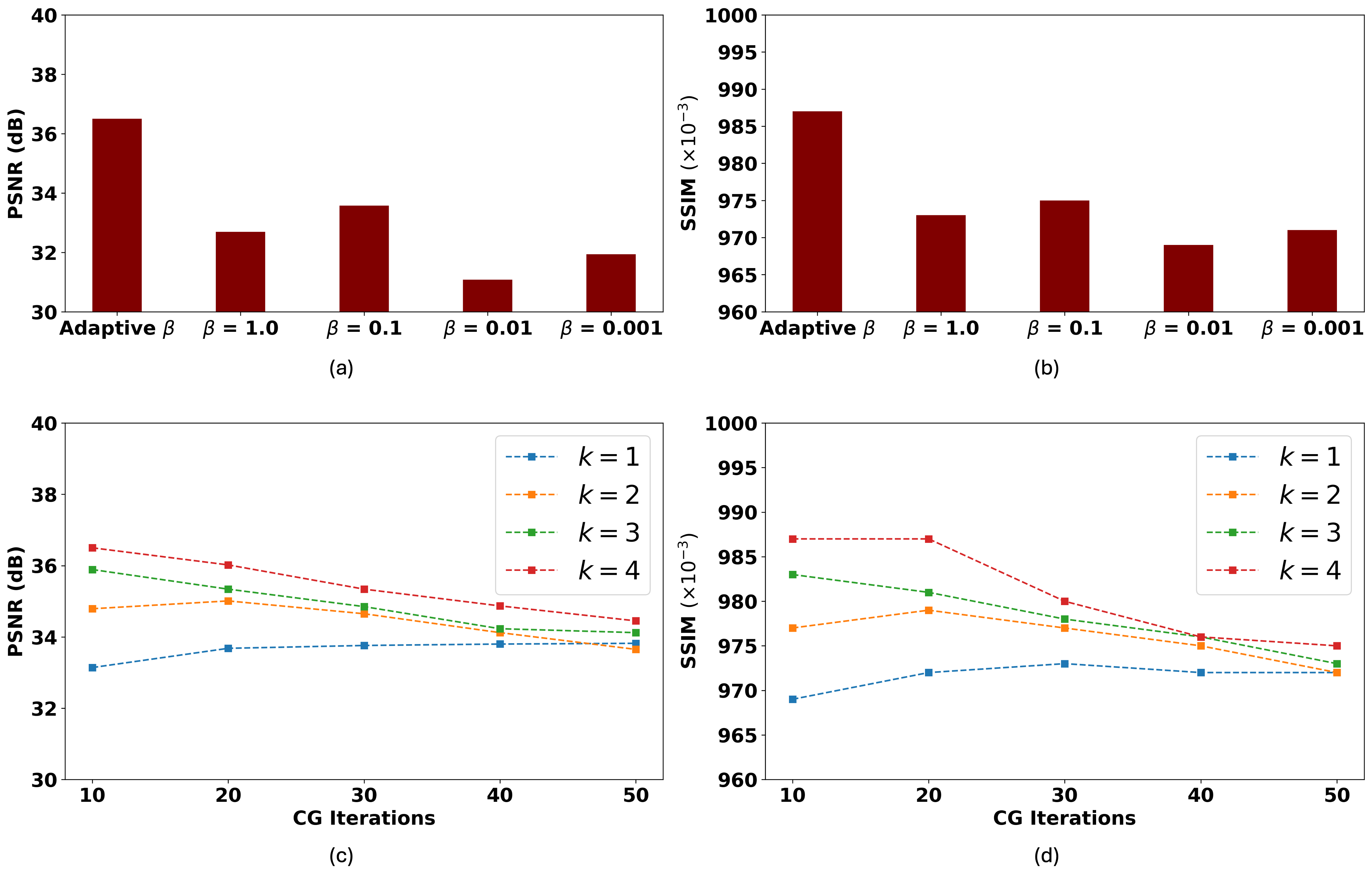}
\caption{Reconstruction performance evaluation of the proposed method in terms of PSNR (in dB) and SSIM on sample ST at 160 kV voltage and 0.6 seconds integration time. Plots (a) and (b) are denoting performance using either fixed or adaptive regularization parameter selection strategy while (c) and (d) represent performance for different iterations of the conjugate gradient algorithm. The curves in (c) and (d) are for the CG in four different outer loops, $k = 1, 2, 3, 4$ respectively.}%
\label{fig:cgb}%
\end{figure*}

% \begin{table*}[ht!]
% \fontsize{9}{12}
% \selectfont
% \centering
% %\renewcommand{\arraystretch}{1.0}
% \begin{tabular}{|ccc|cccccccccc|}
% \hline
% \multicolumn{13}{|c|}{Fixed vs Adaptive Regularization Parameter $\beta$} \\ \hline
% \multirow{2}{*}{Sample} & Source & Integration & \multicolumn{2}{c}{Adaptive $\beta$} & \multicolumn{2}{c}{$\beta = 1.0$} & \multicolumn{2}{c} {$\beta = 0.1$} & \multicolumn{2}{c} {$\beta = 0.01$} & \multicolumn{2}{c|} {$\beta = 0.001$} \\ 
% & Voltage & Time & PSNR & SSIM & PSNR & SSIM & PSNR & SSIM & PSNR & SSIM & PSNR & SSIM\\ \hline 
% ST & 160 kV & 0.6s & \textbf{36.50 dB} & \textbf{0.987} & 32.69 dB & 0.973 & 33.57 dB & 0.975 & 31.08 dB & 0.969 & 31.94 dB & 0.971\\ \hline
% \end{tabular}
% \vspace{1em}
% \caption{Reconstruction performance evaluation in terms of PSNR (in dB) and SSIM on sample ST for the proposed method using either fixed or adaptive regularization parameter selection strategy. The bold entries are for the best results.}
% \label{tab:vb_rt} 
% \end{table*}

%% file: sec/exp.tex
\section{Experiments and Results}
%In this section we report results and discuss about the performance of the proposed algorithm and various state-of-the-art methods. 

% \begin{figure*}[t!]
% 	\centering
% 	\includegraphics[width = \textwidth,keepaspectratio=true,trim={2.2cm 8.3cm 1.8cm 8.5cm},clip]{Figures/mat_acq.pdf}
% 	\caption{Illustration of the performance of different algorithms on objects scanned under different settings than the training data set. 
%     }
% 	\label{fig:mat_acq}
% \end{figure*}

%\subsection{Ablation Study}

% \begin{figure*}[t!]
% 	\centering
% 	\includegraphics[width = \textwidth,keepaspectratio=true,trim={1.3cm 7.1cm 5.2cm 7.4cm},clip]{Figures/vacq.pdf}
% 	\caption{Illustration of the performance of different algorithms on objects scanned under different settings than the training data set. 
%     }
% 	\label{fig:acq}
% \end{figure*}

\begin{figure*}[h!]
	\centering
	\includegraphics[scale=0.6,keepaspectratio=true]{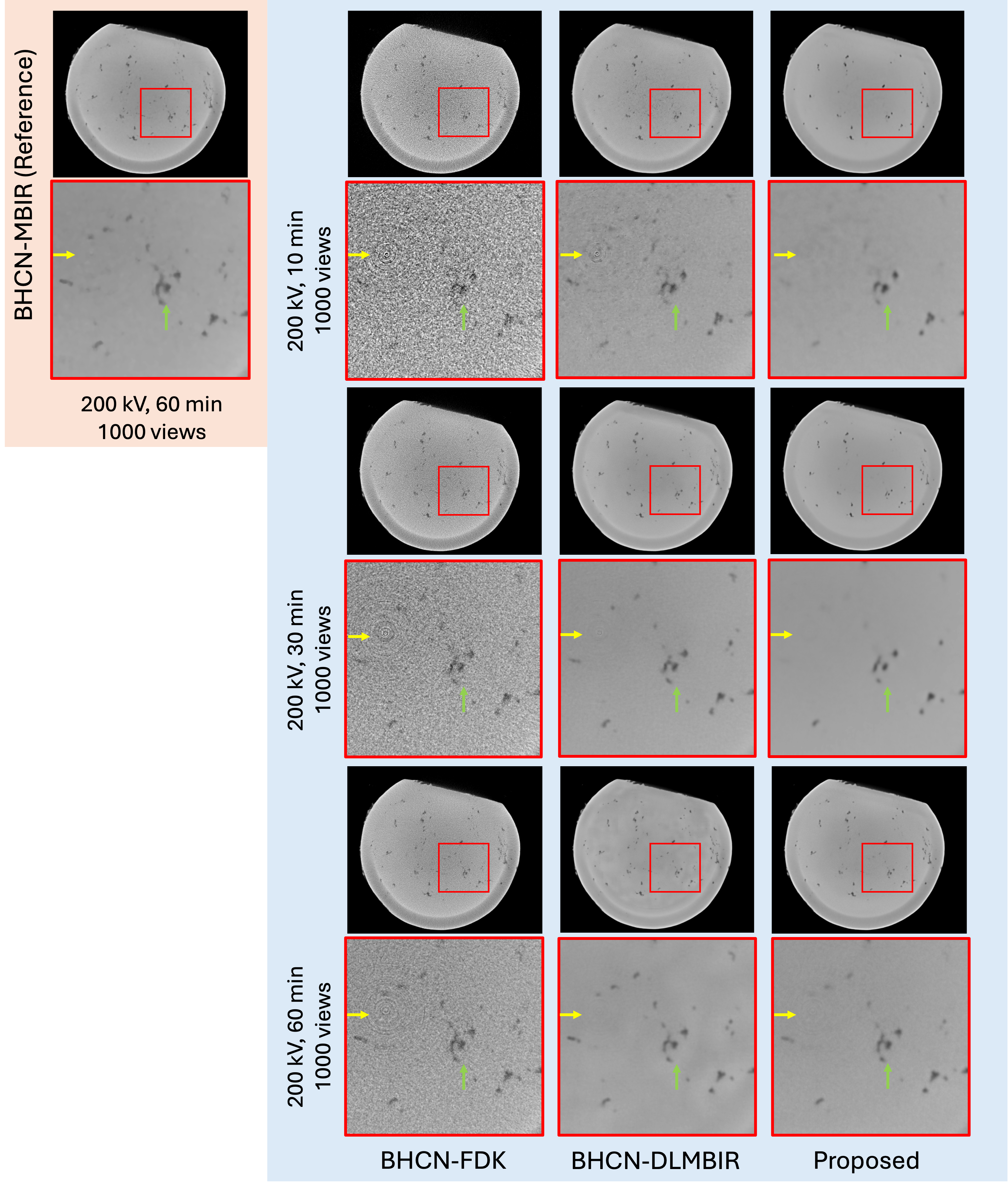}
	\caption{Illustration of the performance on ST sample from Table \ref{tab:ct_data} scanned under different integration times of 0.6 seconds, 1.8 seconds and 3.6 seconds (scan time of 10 minutes, 30 minutes, 60 minutes respectively). The peak source voltage is 200 kV for each case. This is an out-of-distribution data since the training data for DLMBIR and the proposed were from an 8 second integration time. BHCN-MBIR on data from 200 kV, 3.6 seconds integration time (60 minutes scan time) is used as the reference. The yellow arrows point out the ring artifacts that are more prominent in BHCN-FDK and BHCN-DLMBIR reconstructions from lower integration times. The proposed method gets rid of these artifacts for all the integration times shown here. The green arrow shows the pores that have been well preserved by both BHCN-DLMBIR and the proposed method. 
    }
	\label{fig:vint}
\end{figure*}

\begin{figure*}[h!]
	\centering
	\includegraphics[scale=0.6,keepaspectratio=true]{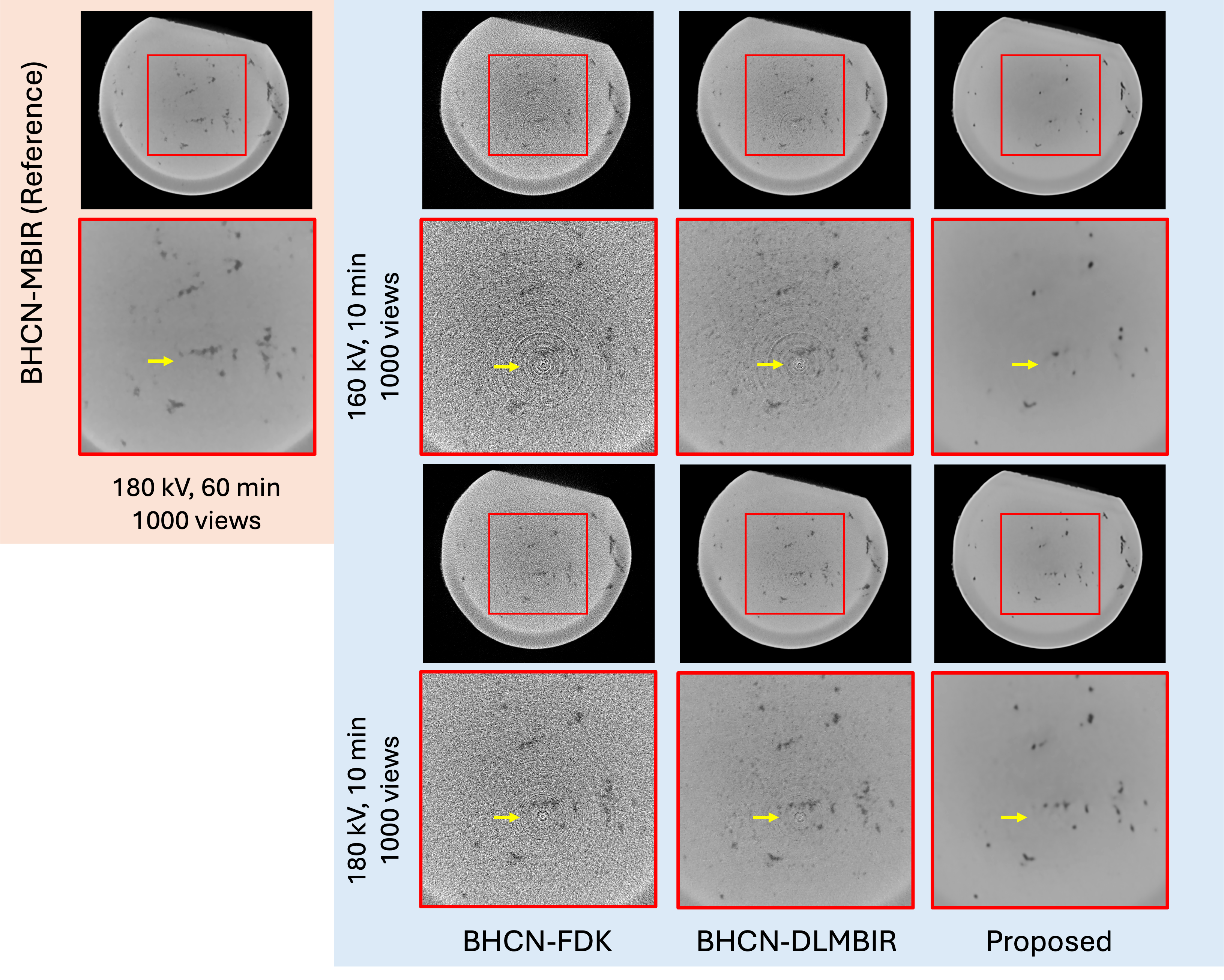}
	\caption{Illustration of the performance of different algorithms on ST sample from Table \ref{tab:ct_data} scanned at 0.6 seconds integration time (total scan time of 10 minutes) for two different X-ray source voltages (160 kV and 180 kV). This is an out-of-distribution data since both BHCN-DLMBIR and the proposed method were trained on data from 200 kV with an integration time of 8 seconds. BHCN-MBIR for 180 kV, 3.6 seconds integration time (60 minutes scan time) is used as the reference since this setting provides significantly less artifacts. The yellow arrows point out the ring artifacts strongly visible in BHCN-FDK as well as BHCN-DLMBIR. The artifact appears to be suppressed in the proposed method. 
    }
	\label{fig:vsv}
\end{figure*}

% \begin{figure*}[t!]
% 	\centering
% 	\includegraphics[width=0.8\textwidth,keepaspectratio=true,trim={1.3cm 7.1cm 11cm 7.4cm},clip]{Figures/vmat.pdf}
% 	\caption{Performance on different materials. 
%     }
% 	\label{fig:mat}
% \end{figure*}

% \begin{figure}[t!]
% 	\centering
% 	\includegraphics[scale=1.2,keepaspectratio=true,trim={1.3cm 4.7cm 13.5cm 5cm},clip]{Figures/vsp_new_1.pdf}
% 	\caption{Performance on different sparsity factors. 
%     }
% 	\label{fig:vsp}
% \end{figure}

\begin{figure*}[h!]
	\centering
	\includegraphics[scale = 0.5,keepaspectratio=true]{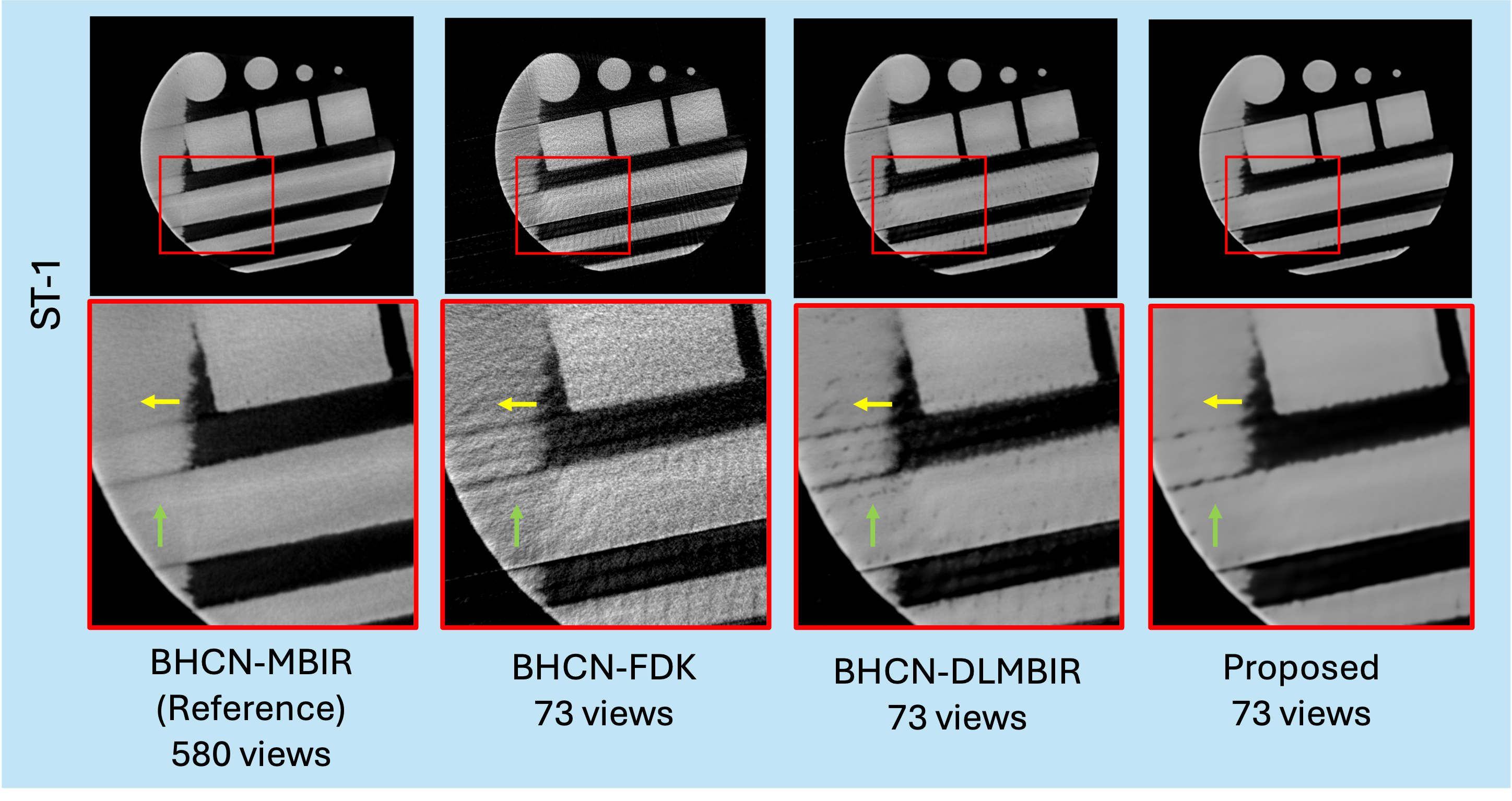}
	\caption{Illustration of performance of various algorithms on ST-1 sample as described in Table \ref{tab:ct_data} at different sparsity factors. This is an out-of-distribution data since the CNNs have been trained with a sparsity factor of four (138 views shortscan) while the methods are being tested on a factor of eight (73 views shortscan). 
    The yellow and green arrows point out the artifacts present in the reconstructions. Both BHCN-FDK and BHCN-DLMBIR show streak artifacts or noise which are suppressed in the proposed reconstruction when compared to the dense-view BHCN-MBIR reconstructions in each case.  
    }
	\label{fig:vsp}
\end{figure*}

The highlight of our algorithm is that it attempts to fulfill all the three criteria for adoption in industries. 
We carefully evaluate these by comparing against the state-of-the-art industrial cone-beam XCT reconstruction methods using metrics such as Peak Signal to Noise Ratio (PSNR) and Structural Similarity (SSIM) \cite{wang2004image}.
We demonstrate differences in cone-beam XCT reconstruction quality through visible artifacts (beam hardening, streak artifacts, noise) and quality of defects such as pores.

\subsection{Performance on Varying Integration Times.}
We tested the performance of the proposed algorithm on steel sample ST scanned at integration times of 0.6 seconds, 1.8 seconds and 3.6 seconds respectively (10 minutes, 30 minutes, 60 minutes scan time) under a source voltage of 200 kV. 
Our algorithm is compared against BHCN-FDK and BHCN-DLMBIR as shown in Fig. \ref{fig:vint} and the performance metrics are reported in Table \ref{tab:metrics} (middle sub-table).
The CNNs are trained for an integration time of 8 seconds and therefore the test data is out-of-distribution.
BHCN-MBIR on ST at 200 kV and 3.6 seconds integration time is used as the reference to compare the reconstructions and compute the metrics.
The FDK reconstructions appear very noisy and contain strong ring artifacts in each case.
The amount of artifacts reduce with increase in integration time.
BHCN-DLMBIR shows some residual ring artifacts as well as noise for the 200 kV, 10 min scan.
Our proposed method provides significantly cleaner images as compared to BHCN-DLMBIR, although it is losing a few smaller pores in the process.
BHCN-DLMBIR performance improves with higher integration times and similar trend is observed for the proposed method as well.
Both the DL methods perform at par for 1.8 seconds (30 minutes scan) and 3.6 seconds (60 minutes scan) integration times.
The differences among the reconstructions are marked by arrows with the yellow ones pointing out the ring artifacts while the green ones show the pore details preserved in each case. 
At 0.6 seconds (low integration time), both FDK and DLMBIR possess ring artifacts whereas the proposed method cleans it up.
The PSNR and SSIM values show similar trend; the proposed method outperforms BHCN-DLMBIR by $\approx 1-1.5$ dB (PSNR) and $\approx 0.01-0.02$ (SSIM).

\subsection{Performance on Varying Source Voltages.}
Our method is compared against BHCN-DLMBIR for varying source voltage (160 kV, 180 kV) while keeping the integration time constant at 0.6 seconds (10 minutes scan) in Fig. \ref{fig:vsv} and the metrics are reported in Table \ref{tab:metrics} (top sub-table).
This is an out-of-distribution setting for both the DL-based methods since the CNNs have been trained on 200 kV data and with longer integration time per view.
The artifacts in BHCN-FDK (rings and noise) are relatively lower at 180kV as compared to 160kV.
BHCN-DLMBIR shows ring artifacts and also appears noisy as pointed by the yellow arrows on the zoomed sections whereas the proposed method gets rid of both.
Although the proposed method loses some minor pores when compared to BHCN-MBIR, the reconstructions overall appear clean when compared to BHCN-DLMBIR.
BHCN-MBIR at 180 kV, 3.6 seconds integration time is used as the reference here.
The PSNR and SSIM metrics reported in the table shows superior performance for the proposed method over BHCN-DLMBIR.
The performance gap narrows down with increase in source voltage and it is negligible for 200 kV (Table \ref{tab:metrics}).

\subsection{Performance on Varying Sparsity Factor.}
We test our algorithm on a sparsity factor different from the one it is trained on and report the results in Fig. \ref{fig:vsp} and the bottom sub-table in Table \ref{tab:metrics} respectively.
The CNNs in both BHCN-DLMBIR and the proposed method were trained on steel samples with 138 views short-scan, which is already very sparse considering the size of detector with 1840 channels.
The methods are tested on sparsely sampled FDK reconstructions of steel samples ST-1 and ST-2 with 73-view shortscan which is another factor of two higher in sparsity and therefore, even more challenging to handle and OOD compared to training data. 
It should be emphasized that 73 views for a 1840 channel detector, is theoretically considered a factor of 25X sub-sampling (considering Nyquist theorem).
The BHCN-FDK reconstruction appears extremely noisy with streak artifacts and lose a lot of pore information.
BHCN-DLMBIR suppresses the artifacts partially and shows some residual streaks as delineated by the yellow and green arrows whereas the proposed method shows fewer streak artifacts.
Our proposed method shows relatively fewer artifacts and appears cleaner than BHCN-DLMBIR.
A dense-view BHCN-MBIR reconstruction is used as the reference to show the differences marked by the arrows in the zoomed sections.
The performance metrics reported for both the samples ST-1 and ST-2 show that the proposed method significantly outperforms BHCN-DLMBIR.

\subsection{Performance on Parts Made of Different Materials}

Material density has a substantial impact on noise texture and distribution in X-ray CT imaging, particularly through the process of beam hardening correction, which introduces additional noise as a known trade-off. 
For lower-density materials like aluminum alloys, the variance of this added noise remains relatively low. 
These materials also require a lower X-ray power for adequate penetration, which leads to reduced scattering, smaller focal spots, and cleaner reconstructions. Conversely, denser materials, such as Inconel 718, face increased beam hardening effects and need higher X-ray power to achieve penetration. 
The higher power leads to larger spot sizes, increased scattering, and higher noise variance, all of which contribute to more complex reconstruction challenges. Consequently, models trained on stainless steel could struggle to generalize effectively to materials with different densities, such as aluminum alloy and Inconel, due to the distinct noise characteristics introduced by these density-related variations.

\begin{figure}[h!]
	\centering
	\includegraphics[scale = 1.1,keepaspectratio=true,trim={1.3cm 5.9cm 13cm 6.3cm},clip]{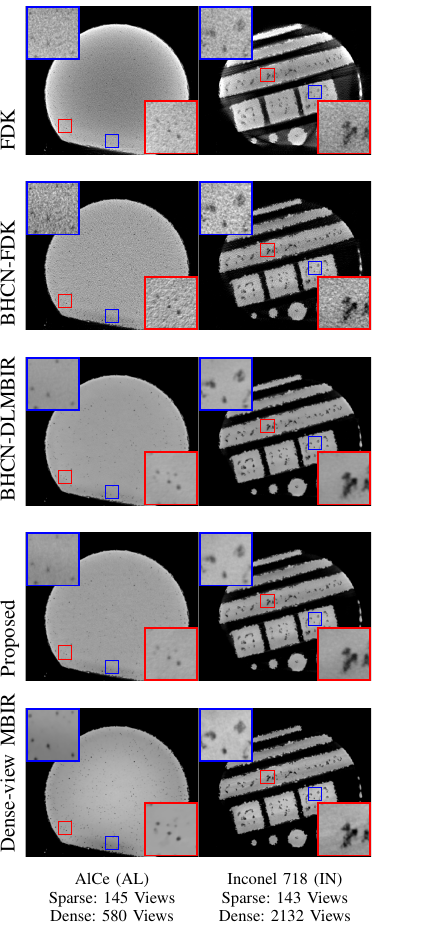}
	\caption{Reconstruction results on parts made of AlCe and Inconel 718. This is out-of-distribution data since the DL methods have been trained on steel data only. All the reconstructions are on sparse views except the dense-view MBIR that is considered as the ground truth. The pores and edge details are well preserved by both BHCN-DLMBIR and the proposed method as shown in the zoomed images.
    }
	\label{fig:mat}
\end{figure}

We evaluate performance on parts manufactured with with out-of-distribution materials ALC (AlCe, Aluminium Cerium, density $\thickapprox  3.4 gr/cm^3$) and IN (Inconel 718, density $\thickapprox 8.2 gr/cm^3$).
The part geometry is similar to the steel samples (ST, ST-1, ST-2).
The results are summarized in Fig. \ref{fig:mat} and Table \ref{tab:mmat}, respectively.

% The DL methods were trained with steel data with different pore and pixel intensity distributions and therefore these are out-of-distribution datasets.
The reconstruction quality is compared against the corresponding dense-view MBIR \cite{yu2010fast} as ground truth.
The sparse views are 145 and 143 while the corresponding dense-view are 580 and 2132 for AlCe (ALC) and Inconel 718 (IN) respectively.
The zoomed regions show clean and sharp edges with the pore details well preserved in both BHCN-DLMBIR and the proposed method.
Some of the minor pores appear smeared in both the methods.
The FDK reconstructions appear extremely noisy.
In terms of metrics, the proposed method performs at par with BHCN-DLMBIR as shown in Table \ref{tab:mmat}.

\begin{table*}[ht!]
\fontsize{8}{11}
\selectfont
\centering
\begin{tabular}{|c|cccccc|}
\hline
\multicolumn{7}{|c|}{Variation in Materials} \\ \hline
\multirow{2}{*}{Sample} & \multicolumn{2}{c}{BHCN-FDK} & \multicolumn{2}{c}{BHCN-DLMBIR} & \multicolumn{2}{c|} {Proposed}\\ 
& PSNR $(\uparrow)$ & SSIM $(\uparrow)$ & PSNR $(\uparrow)$ & SSIM $(\uparrow)$ & PSNR $(\uparrow)$ & SSIM $(\uparrow)$\\ \hline 
ALC & 25.33 dB & 0.412 & \textbf{35.45 dB} & \textbf{0.967} & 35.27 dB & 0.966 \\ 
IN & 29.84 dB & 0.548 & \textbf{37.63 dB} & \textbf{0.976} & 37.56 dB & 0.975 \\ \hline
\end{tabular}
\vspace{1em}
\caption{Performance evaluation in terms of PSNR (in dB) and SSIM for parts manufactured with Aluminium Cerium Alloy (ALC) and Inconel 718 (IN). 
The metrics are computed against dense-view MBIR \cite{yu2010fast} as ground truth.
The sparsity factor (number of views) is in the range similar to that of the training data.
The proposed method performs at par with BHCN-DLMBIR.
The best metric values for each case are highlighted in bold.
}
\label{tab:mmat} 
\end{table*}

\subsection{Adaptive vs Fixed Regularization Parameter.}
\begin{figure*}[h!]
	\centering
	\includegraphics[width=\textwidth,keepaspectratio=true]{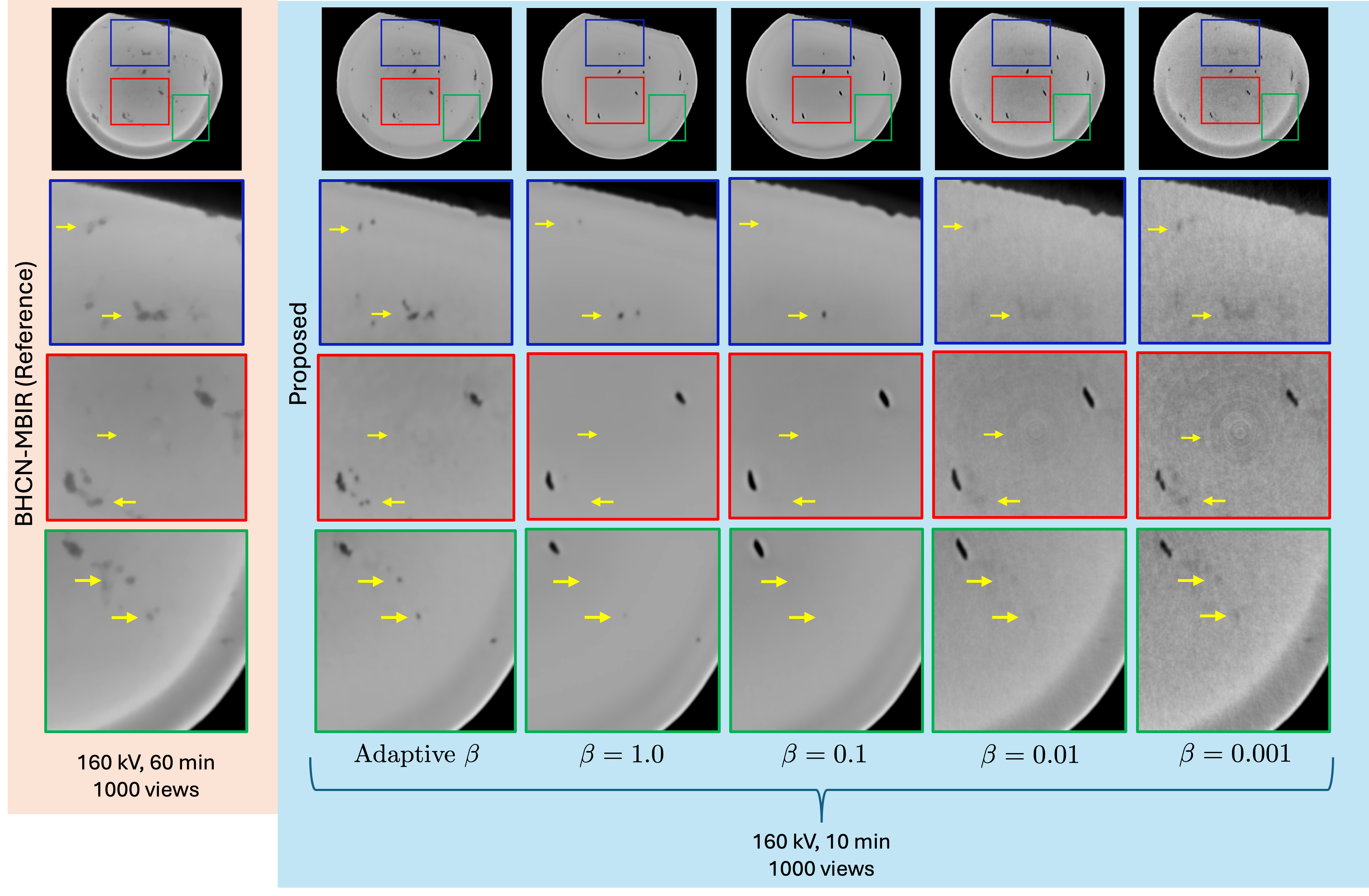}
	\caption{Performance comparison of the proposed method for adaptive regularization parameter $\beta$ against a fixed $\beta$ across the outer iterations. The sample is ST scanned at 160 kV source voltage with a 0.6 second integration time (10 minutes total scan time) as described in Table \ref{tab:ct_data}. BHCN-MBIR reconstruction on ST sample from a 160 kV, 3.6 seconds integration time scan is used as the reference. The yellow arrows indicate the pores missing from reconstructions corresponding to $\beta = 1.0, 0.1, 0.01, 0.001$. These details are well preserved in the proposed reconstruction with adaptive $\beta$ selection strategy introduced in this work.}
	\label{fig:fva}
\end{figure*}

The proposed algorithm uses an automatic regularization parameter selection algorithm at every outer iteration $k$ whereas traditional iterative algorithms keep the parameter fixed.
In Fig. \ref{fig:fva} and \ref{fig:cgb} (a), (b), we compare the reconstruction quality of our method with adaptive regularization against fixed regularization $\beta$ for $\beta = \{1.0, 0.1, 0.01, 0.001\}$ to show the benefit of our automated parameter selection strategy.
The reconstructions are from ST sample described in Table \ref{tab:ct_data} at 160 kV and 0.6 seconds integration time.
BHCN-MBIR reconstruction of ST at 180 kV and 3.6 seconds integration time is used as the reference for comparisons.
The pores are better preserved with the adaptive $\beta$ strategy as pointed by the yellow arrows. 
The reconstruction from $\beta = 1.0$ weighs the CNN more and therefore suffers from excessive blurring.
From $\beta = 0.1$ onwards weight to the regularization term decreases and therefore the streak and noise due to the scan settings become more prominent. 
The pores are lost for lower $\beta$ values.
Overall, the adaptive $\beta$ shows improved reconstruction quality over fixed ones.
The bar plots in Fig. \ref{fig:cgb} (a), (b) quantifies the reconstruction performance for different values of parameter $\beta$ for the sample ST at 160 kV voltage and 0.6 seconds integration time.
The ground truth for calculation of metrics was BHCN-MBIR reconstruction of sample ST at 180 kV and 3.6 seconds integration time.
The adaptive strategy outperforms the fixed $\beta$ in terms of PSNR and SSIM as shown in Fig. \ref{fig:cgb} (a), (b).

\subsection{Variation in Performance with CG Iterations}
Another parameter in our proposed algorithm is the number of iterations in the CG step solving for \eqref{vimg}.
We keep the number of CG iterations fixed across the outer loop and study their effect on reconstruction quality.
The experiments were performed on the sample ST at 160 kV voltage and 0.6 seconds integration time.
The algorithm was run for $K = 4$ outer iterations and for a different number of CG iterations: $\{10, 20, 30, 40 ,50\}$.
For each outer loop $k \leq K$ we plot the PSNR and SSIM metrics of the reconstruction obtained from the CG step in Fig. \ref{fig:cgb} (c), (d).
The metrics were calculated against BHCN-MBIR reocnstruction on the sample ST at 180 kV voltage and 3.6 seconds integration time.
For initial two outer iterations ($k \leq 2$), there is a slight improvement in PSNR performance with increase in iterations from 10 to 20 and it drops slightly or remains stable after 30 iterations as shown by the blue and orange curves.
The green and red curves for $k = 3$ and $k = 4$, respectively, show drop in PSNR performance for iterations greater than 10 which implies optimal performance has already been achieved at the 10 CG iterations mark.
A similar observation is found for SSIM as well.
Since this is an automatic regularization parameter selection strategy for $\beta$, the drop or improvement in performance with CG iterations at a specific outer iteration $k$ depends on the chosen $\beta$ value as well.

\subsection{Computational Complexity Comparisons.}
A key benefit of the proposed approach is the reduced run-time over MBIR. 
MBIR has been the state-of-the-art technique for industrial XCT reconstruction, however, the run-time is high and undesirable for larger data. 
We report average run-times of various methods in Table \ref{tab:runtime} for cone-beam XCT data described in Table \ref{tab:ct_data} when using a system with CPU and four NVIDIA P100 GPUs. 
Each algorithm is designed to utilize the multi-core CPU and GPUs. 
The dimensions of the reconstructed 3D volumes are approximately $1500 \times 1800 \times 1800$. 
MBIR on $\approx 150$ views takes $\approx$ 27000 seconds (7 hours 30 minutes for 120 iterations) on average which is significantly slower than the scan time ($10-20$ minutes) when using the pyMBIR \cite{venkat2019pymbir} library. 
The proposed algorithm requires three or fewer iterations ($\leq$ 1600 seconds) and is therefore faster than MBIR, reducing the run-time by an order of magnitude and making it comparable to the acquisition times. 
This illustrates the ability of our method to be utilized as an in-line tool for industrial XCT systems in practice.  
The CNN prior provides very fast computations on GPUs compared to the Markov Random Field prior used by MBIR. 
A major chunk of time is consumed by the CG block of our algorithm. 
Both FDK and DLMBIR show reduced run-time by an order of magnitude when compared to the iterative methods.

\begin{table}[h!]
\fontsize{8}{11}
\selectfont
\centering
\begin{tabular}{|cccc|}
\hline
\multicolumn{4}{|c|}{Average Run time per reconstruction in seconds} \\
\hline
MBIR & DLMBIR & Proposed & FDK\\  \hline 
27132 s & 325 s & 1612 s & 75 s \\ \hline
\end{tabular}
\vspace{1em}
\caption{Runtime Comparisons}
\label{tab:runtime} 
\end{table}

% \begin{figure*}[h!]
% 	\centering
% 	\includegraphics[width=0.8\textwidth,keepaspectratio=true]{Figures/var_cg.png}
% 	\caption{Performance comparisons of the proposed method for different number of iterations of CG (10, 30, 50) per outer iteration. The sample is ST scanned at 160 kV source voltage with a 0.6 second integration time (10 minutes total scan time) as described in Table \ref{tab:ct_data}. BHCN-MBIR reconstruction on ST sample from a 160 kV, 3.6 seconds integration time scan is used as the reference.}
% 	\label{fig:vcg}
% \end{figure*}

%% file: sec/con.tex
\section{Conclusion}

In the context of industrial XCT, current methods have different limitations in terms of computational complexity, GPU memory demand and reconstruction performance. 
We address these challenges by introducing a generic DL regularized iterative optimization algorithm.
The proposed method uses a pre-trained artifact removal CNN to regularize the image recovery. 
We employ regularization parameter selection at every iteration to improve performance due to the varying noise statistics of the reconstructions from successive iterations.
The designed algorithm is general enough to be implemented for any other inverse imaging problem. 
The proposed algorithm has been validated on real cone-beam XCT scans of various 3D printed materials. 
We demonstrate that our method shows better generalizability over single-step DL for variation in scanning parameters (X-ray source voltage, integration time and sparsity factor).
The algorithm reduces computational complexity significantly as compared to MBIR.
Finally, the automated selection of regularization parameter to produce reasonable visual quality across different imaging conditions helps overcome a prevalent challenge with current iterative reconstruction algorithms. 
One of the limitations of the proposed approach may be losing smaller features and pores in the reconstruction. 
This could be critical in some applications, and therefore, our future research will focus on how to balance the loss function to ensure the resolving power of the proposed approach remains high while producing high quality reconstruction for in- and out-of-distribution data.